\documentclass[sigconf]{acmart}

\renewcommand\footnotetextcopyrightpermission[1]{}

\usepackage{booktabs}
\usepackage{amsmath}
\usepackage{tikz}
\usetikzlibrary{arrows.meta, positioning}
\usepackage{listings}

\begin{document}

\title{ How Often Should a Recommender Call an LLM?  Value-Weighted Routing, Monitoring, and Seasonal Robustness }

\author{Bhavtosh Rath}
\affiliation{%
  \country{USA} 
}
\email{rathbhavtosh3003@gmail.com}

\renewcommand{\shortauthors}{Rath}

\begin{abstract}
Routing decisions between a cheap heuristic and an expensive large-language-model (LLM) call is usually framed as a difficulty problem: send the hard cases to the expensive path. We argue this framing is incomplete, because difficulty and business value are separate axes -- a difficult four-dollar item and a difficult two-thousand-dollar item are not the same mistake to get wrong. We present \textit{value-router}, a fully synthetic simulation study of a retail merchandising pipeline that routes items to a fast or slow decision path using both an estimated difficulty and an estimated value, never ground truth. The system is built and evaluated in three stages. First, a value-weighted threshold router is compared against a difficulty-only baseline and a random baseline on a synthetic catalog with a deliberate inverse correlation between category volume and category value; value-weighting matches the difficulty-only baseline's recall of true high-value items ($\approx$60\%) while achieving substantially higher precision (98.3\% vs.\ 94.3\%), i.e., it wastes less of the expensive path on hard-but-cheap items. Second, a decision logger and monitor reveal a failure mode invisible to aggregate metrics: a difficulty estimator with a respectable overall estimate-truth correlation (0.792) collapses to near zero once measured \textit{within} each catalog category, showing the aggregate number was almost entirely a between-category effect rather than genuine per-item discriminative power. Third, a simulated Black-Friday-style demand surge (2.5$\times$ volume, a mix shift toward higher-value categories) is used to compare a season-blind static router, a router with a manually calibrated seasonal threshold, and two slow-path budget policies. A naive fixed daily budget collapses high-value recall from 70.1\% to 16.2\% under the surge, while an elastic budget that simply scales with that day's own value estimates -- with no knowledge that a peak is occurring -- tracks the surge automatically and matches the unconstrained router almost exactly. All results are from a controlled synthetic simulation with experimenter-defined ground truth and are intended to demonstrate design principles for cost-aware routing systems, not to constitute a validated real-world claim.
\end{abstract}


\maketitle

\section{Introduction}

As large language models become embedded in personalization and decisioning pipelines, every additional LLM call carries a real cost in latency, tokens, and compute. This raises a basic systems question: \textit{when should a system spend an expensive reasoning call, and when is a cheap heuristic just as good?}

A common first answer is ``spend more reasoning on harder cases.'' This is an incomplete answer. Consider two items entering a retail catalog: a \$4 phone case that is genuinely ambiguous to categorize, and a \$2{,}000 leather jacket that is equally ambiguous. Both are \textit{difficult} in the same sense -- a simple rule could plausibly get either one wrong -- but a wrong decision on the jacket is worth far more than a wrong decision on the phone case. Difficulty tells us how uncertain a decision is; value tells us how costly a mistake would be. A routing policy that only looks at difficulty cannot distinguish these two cases, and will spend expensive reasoning on cheap-but-hard items exactly as often as it spends it on expensive-but-hard ones.

This paper presents \textit{value-router}, a simulation study built around that distinction, developed in three stages that were each designed to be independently useful:

\begin{itemize}
  \item \textbf{Core routing (Section~\ref{sec:tier1})}: a synthetic catalog simulator with observable fields (category, price) and hidden ground-truth fields (value, difficulty); a difficulty scorer and a value estimator that only ever see the observable fields; and a router that sends an item to the expensive path only if it clears \textit{both} an estimated-difficulty and an estimated-value threshold. This is compared against a difficulty-only baseline and a random baseline.
  \item \textbf{Monitoring (Section~\ref{sec:tier2})}: a decision logger and a monitor that check two things an aggregate metric can hide -- whether routing spend is concentrated on the rare, high-value segment as intended, and whether the estimators feeding the router are actually well calibrated \textit{within} each category, not just across the whole catalog.
  \item \textbf{Seasonal robustness (Section~\ref{sec:tier3})}: a multi-day simulation of a demand surge (a stand-in for a real seasonal event such as Black Friday), used to compare a season-blind router, a manually calibrated calendar-aware router, and two ways of capping the expensive path's budget -- one fixed, one elastic to the day's own estimates.
\end{itemize}

We want to be explicit about scope: every number in this paper comes from a fully synthetic simulation with experimenter-defined ground truth, run at a single random seed per configuration unless otherwise noted, over a system built by one person as a design exercise. The contribution is a set of design patterns and the specific, sometimes counter-intuitive findings that emerged from testing them against ground truth we deliberately hid from the system under test -- not a benchmark result that should be expected to transfer directly to a production deployment. Section~\ref{sec:discussion} returns to this point.

\section{Related Work}
\label{sec:related}

Reducing the cost of LLM-heavy pipelines by routing only a subset of inputs to an expensive model is an active area of work. FrugalGPT~\cite{chen2023frugalgpt} learns a cascade over multiple LLM APIs of increasing cost, escalating a query only when a cheaper model's answer is judged insufficient, building on the earlier FrugalML~\cite{chen2020frugalml} formulation of cascading across paid prediction APIs under a budget constraint. LLM cascades built around intermediate reasoning representations~\cite{yue2023cascades} and mixture-of-model approaches such as AutoMix~\cite{madaan2023automix} pursue a similar goal: spend the expensive model's reasoning only where a cheaper stage's output is not trustworthy. Model cascading has also been studied more generally as a way to jointly improve efficiency and accuracy in NLP systems~\cite{varshney2022model}, and system-level realizations such as Tabi~\cite{wang2023tabi} show that multi-level cascades deliver measurable latency wins in deployed LLM-serving infrastructure. RouteLLM~\cite{ong2024routellm} and Hybrid LLM~\cite{ding2024hybrid} instead learn a router directly from preference or quality-labeled data, framing routing as a learned classification problem over query features, and EcoAssistant~\cite{zhang2023ecoassistant} applies a similar cost-quality escalation strategy to tool-augmented LLM assistants.

A second relevant thread is confidence calibration, since most of the cascades above escalate based on some notion of the cheap path's uncertainty. Classic work on neural network calibration~\cite{guo2017calibration} established that aggregate calibration metrics can mask systematic miscalibration within subpopulations of the data -- the same failure mode we surface for our difficulty estimator in Section~\ref{sec:tier2}, here in an LLM-routing rather than image-classification setting. More recent work asks whether LLMs themselves know what they don't know, either via internal probability estimates~\cite{kadavath2022language} or via elicited verbalized confidence~\cite{tian2023just}; both lines of work are natural sources of the difficulty/uncertainty signal that a cascade router such as ours would consume in a non-synthetic deployment. Cost-sensitive learning more broadly -- weighting errors by their real-world cost rather than treating all misclassifications equally~\cite{elkan2001foundations} -- is the classical decision-theoretic framing underlying our value axis, applied here to inference-time routing rather than model training.

Finally, we distinguish this line of work from techniques that reduce the cost of a single expensive call rather than deciding whether to make one at all: speculative decoding~\cite{leviathan2023fast} accelerates autoregressive generation from a large model using a smaller draft model, and sparsely-activated architectures such as Switch Transformers~\cite{fedus2022switch} route each token to a subset of a single model's own parameters. Both are complementary to, rather than competing with, an external cascade router: a deployment could route between a fast and slow \textit{path} using the mechanisms surveyed above, while each path internally uses speculative decoding or sparse routing to further cut cost.

This body of work is almost entirely organized around a single axis: is the cheap path's output good enough, or does this query need the expensive path? \textit{value-router} asks a related but distinct question, motivated by a merchandising/e-commerce setting rather than open-domain query answering: even restricting attention to inputs where a cheap heuristic is genuinely likely to be wrong, is it worth paying for the expensive path, given what a mistake on this particular input would cost the business? The two questions are complementary -- a production system could plausibly use a learned cascade router \textit{and} a value gate on top of it -- but the routing literature above does not, to our knowledge, treat the cost of being wrong as a first-class second axis alongside the difficulty of getting it right. \textit{value-router} also differs in evaluation methodology: because the environment is fully synthetic, every routing decision can be graded against experimenter-defined ground truth that a router is never shown, which is what makes the calibration analysis in Section~\ref{sec:tier2} and the demand-surge analysis in Section~\ref{sec:tier3} possible without needing deployed traffic.

Taken together with the systems surveyed above, we see \textit{value-router}'s contribution as three specific, testable ideas that do not appear jointly in the cited work:

\begin{itemize}
  \item \textbf{A value axis orthogonal to the uncertainty axis.} Cascade and routing systems~\cite{chen2023frugalgpt, chen2020frugalml, yue2023cascades, madaan2023automix, ong2024routellm, ding2024hybrid, varshney2022model, wang2023tabi, zhang2023ecoassistant} and calibration/confidence-elicitation work~\cite{guo2017calibration, kadavath2022language, tian2023just} both operate on a single scalar -- is the cheap path good enough? Cost-sensitive learning~\cite{elkan2001foundations} weights \textit{training} errors by cost but has not, to our knowledge, been combined with an uncertainty-based escalation decision at \textit{inference} time. Section~\ref{sec:tier1} shows this combination is not redundant: on a catalog with realistic volume/value structure, adding a value gate to an uncertainty-only router improves precision by 4 points at equal recall, i.e., value is not merely a proxy for difficulty even when the two happen to be correlated.
  \item \textbf{Within-segment calibration as a routing diagnostic, not just a training-time check.} The calibration literature~\cite{guo2017calibration} establishes that aggregate calibration can hide subgroup miscalibration; we show this is not a theoretical concern but the dominant failure mode of the specific estimator feeding a routing decision (0.792 aggregate correlation collapsing to $\approx$0 within every category, Section~\ref{sec:tier2}), and we frame it as a monitoring check a deployed router should run continuously against its own segments, not a one-time audit performed before deployment.
  \item \textbf{A budget policy that adapts to load without being told a season occurred.} Where prior systems either fix a cost budget or learn a router offline, the elastic budget in Section~\ref{sec:tier3} caps spend as a fraction of the router's own running value estimates rather than a fixed figure, and empirically absorbs a 2.5$\times$ demand surge with zero season-specific logic -- outperforming a router given explicit calendar knowledge (Table~\ref{tab:peak}). We are not aware of prior cascade or routing work that evaluates budget policies under a non-stationary traffic shift of this kind.
\end{itemize}

Each of these three ideas is deliberately simple -- a second threshold, a groupby, a ratio -- which we view as a feature rather than a limitation: the paper's claim is that value-awareness, segment-conditioned monitoring, and load-adaptive budgeting are underused design primitives in current LLM routing systems, not that they require new modeling machinery to realize.

\section{Core Routing System}
\label{sec:tier1}

\subsection{Catalog simulator}

The simulator generates items across five categories -- \texttt{commodity}, \texttt{accessory}, \texttt{mid\_tier}, \texttt{premium}, \texttt{luxury} -- each with its own volume weight and per-category ranges for price and margin. For an item $i$ with sampled price $p_i$ and margin $m_i$, ground-truth value is defined deterministically as

\begin{equation}
  \text{value}_i = p_i \cdot m_i .
\end{equation}

Difficulty $d_i \in [0,1]$ is sampled independently of value within a category-specific range, standing in for however hard a real merchandising decision on that item would be to get right -- it does not model any specific real-world mechanism, only the fact that some items are harder to decide on than others.

The category configuration deliberately encodes an \textit{inverse} correlation between volume and value (Table~\ref{tab:catalog}): the highest-traffic category (\texttt{commodity}) has the lowest per-item value, and the rarest category (\texttt{luxury}) has the highest. This mirrors a common e-commerce shape and is the condition that stresses a naive router: a policy optimizing for ``handle the most items well'' will gravitate toward the high-volume, low-value categories and can silently under-serve the rare, high-value tail. A \texttt{--no-inverse} control flag re-weights all categories equally, removing this effect for comparison.

\begin{table}[t]
  \centering
  \caption{Default synthetic catalog, $n{=}2{,}000$, seed 42. \texttt{luxury} is 3.1\% of volume but $\sim$700$\times$ the mean value of \texttt{commodity}.}
  \label{tab:catalog}
  \small
  \begin{tabular}{lrrr}
    \toprule
    Category & Volume \% & Mean value & Mean price \\
    \midrule
    commodity & 39.5 & 0.77   & 8.95 \\
    accessory & 32.4 & 3.83   & 25.50 \\
    mid\_tier & 17.2 & 20.48  & 95.56 \\
    premium   & 7.8  & 92.92  & 323.36 \\
    luxury    & 3.1  & 552.44 & 1{,}391.97 \\
    \bottomrule
  \end{tabular}
\end{table}

\subsection{Difficulty scorer and value estimator}

The router never sees value or difficulty directly. Instead, two estimators produce approximations from the observable fields alone (category, price):

\begin{align}
  \widehat{d}_i &= \text{baseline}(c_i) + \beta(c_i)\cdot\frac{p_i}{500} + \varepsilon_d, & \varepsilon_d &\sim \mathcal{N}(0, 0.05) \\
  \widehat{v}_i &= p_i \cdot \big(\bar{m}(c_i) + \varepsilon_v\big), & \varepsilon_v &\sim \mathcal{N}(0, 0.02)
\end{align}

where $\text{baseline}(c_i)$ and $\beta(c_i)$ are a per-category difficulty baseline and price sensitivity, and $\bar{m}(c_i)$ is a per-category \textit{expected} margin -- a business would plausibly have this kind of historical, category-level average on hand, but not an individual item's exact margin before the fact. Gaussian noise on top represents the imprecision of a real estimator. Both estimators are evaluated against the simulator's hidden ground truth purely for grading; the router itself only ever consumes $\widehat{d}_i$ and $\widehat{v}_i$.

\subsection{Router}

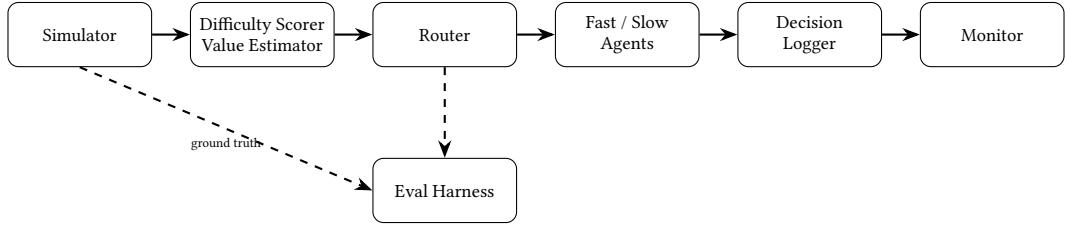
\begin{figure*}[t]
  \centering
  \begin{tikzpicture}[
    box/.style={draw, rectangle, rounded corners, align=center, minimum height=0.85cm, minimum width=1.9cm, font=\footnotesize, inner sep=2pt},
    arrow/.style={-{Stealth}, thick},
    node distance=0.5cm
  ]
    \node[box] (sim) {Simulator};
    \node[box, right=of sim] (est) {Difficulty Scorer\\Value Estimator};
    \node[box, right=of est] (router) {Router};
    \node[box, right=of router] (agents) {Fast / Slow\\Agents};
    \node[box, right=of agents] (log) {Decision\\Logger};
    \node[box, right=of log] (mon) {Monitor};
    \node[box, below=1.2cm of router] (eval) {Eval Harness};

    \draw[arrow] (sim) -- (est);
    \draw[arrow] (est) -- (router);
    \draw[arrow] (router) -- (agents);
    \draw[arrow] (agents) -- (log);
    \draw[arrow] (log) -- (mon);
    \draw[arrow, dashed] (sim.south) -- (eval.west) node[midway, below, font=\tiny] {ground truth};
    \draw[arrow, dashed] (router.south) -- (eval.north);
  \end{tikzpicture}
  \caption{The core pipeline (Tiers 1--2). Only \texttt{category} and \texttt{price} are observable to the router; \texttt{value} and \texttt{difficulty} are hidden ground truth consumed only by the Eval Harness and Monitor, never by the router itself.}
  \label{fig:pipeline}
\end{figure*}

The primary router, \textsc{ValueWeighted}, sends an item to the expensive path only if it clears \textit{both} thresholds:

\begin{equation}
  \text{path}(i) =
  \begin{cases}
    \text{slow} & \text{if } \widehat{d}_i \ge \tau_d \ \text{and} \ \widehat{v}_i \ge \tau_v \\
    \text{fast} & \text{otherwise}
  \end{cases}
  \label{eq:route}
\end{equation}

with defaults $\tau_d{=}0.5$, $\tau_v{=}20$. Two baselines isolate the effect of dropping the value signal: \textsc{DifficultyOnly} uses only the $\widehat{d}_i \ge \tau_d$ condition, and \textsc{Random} sends each item to the slow path independently with fixed probability, ignoring both estimates. Figure~\ref{fig:pipeline} shows the full pipeline, including the downstream fast/slow agents (Section~\ref{sec:agents}) and the evaluation/monitoring components introduced in Sections~\ref{sec:tier1eval} and~\ref{sec:tier2}.

\subsection{Fast and slow path agents}
\label{sec:agents}

Both paths ultimately produce one of five merchandising actions -- \textit{feature}, \textit{standard}, \textit{discount}, \textit{bundle}, \textit{deprioritize} -- a business-rule layer that would sit between a page's base ranking/recommendation system and the final rendered page (e.g.\ \textit{feature} boosts an item's rank or gives it a promoted slot; \textit{deprioritize} suppresses it from promoted placement). The fast path is a flat category-to-action lookup table. The slow path has two interchangeable implementations behind the same interface: a deterministic mock that follows a small decision rubric over $(\widehat{d}_i, \widehat{v}_i)$ and emits a templated one-sentence rationale, and a live implementation that calls the Claude API with a structured-output schema constraining the response to the fixed action set, for cases where a real reasoning call is desired instead of the free deterministic stand-in.

\subsection{Evaluation methodology}
\label{sec:tier1eval}

Every routing decision costs $c_{\text{fast}}{=}1$ or $c_{\text{slow}}{=}20$ unit-cost, an arbitrary but fixed ratio standing in for ``cheap heuristic'' vs.\ ``expensive LLM call'' -- only the ratio matters, not the absolute numbers. Let $H$ be the set of items with true value above $\tau_v$ (``true high-value''), and $S$ be the set of items a policy actually routes to the slow path. We report:

\begin{align}
  \text{recall} &= |H \cap S| / |H| \\
  \text{precision} &= |H \cap S| / |S| \\
  \text{efficiency} &= \sum_{i \in S}\text{value}_i \, \Big/ \sum_{i} c_{\text{path}(i)}
\end{align}

Recall asks: of the items that truly deserved careful treatment, how many got it? Precision asks: of the items that got careful treatment, how many actually deserved it? Efficiency is true value protected per unit of budget spent.

\subsection{Results: value-weighting vs.\ baselines}
\label{sec:tier1results}

Table~\ref{tab:tier1} compares the three strategies on the default catalog ($n{=}2{,}000$, seed 42). \textsc{ValueWeighted} and \textsc{DifficultyOnly} achieve nearly identical recall ($\approx$60\%) -- unsurprising on this catalog, since difficulty and value are positively correlated across categories by construction (see Table~\ref{tab:catalog}), so difficulty alone is already a fair value proxy here. The strategies diverge on precision: \textsc{ValueWeighted} wastes noticeably less slow-path budget on hard-but-not-valuable items (98.3\% vs.\ 94.3\%). Both crush \textsc{Random} on efficiency by a factor of $\sim$5.5--5.6$\times$, confirming that either signal alone is far better than no signal at all.

\begin{table}[t]
  \centering
  \caption{Tier 1: strategy comparison, $n{=}2{,}000$, seed 42. Budget and value are in unit-cost / value units, not currency.}
  \label{tab:tier1}
  \small
  \begin{tabular}{lrrrrr}
    \toprule
    Strategy & Slow \% & Budget & Recall & Precision & Efficiency \\
    \midrule
    ValueWeighted   & 11.8 & 6{,}503 & 60.4\% & 98.3\% & 7.666 \\
    DifficultyOnly  & 12.4 & 6{,}712 & 60.6\% & 94.3\% & 7.451 \\
    Random          & 20.8 & 9{,}923 & 20.5\% & 18.9\% & 1.367 \\
    \bottomrule
  \end{tabular}
\end{table}

A per-category starvation check -- the correlation between a category's share of total volume and that category's own high-value recall -- is strongly negative for all three strategies ($-0.944$, $-0.944$, $-0.976$ respectively), i.e.\ low-volume/high-value categories are not being systematically under-covered relative to high-volume ones on this catalog. This is a property of the specific synthetic configuration (difficulty and value happen to be positively category-correlated by default) rather than a general guarantee, which is exactly the kind of segment-level question Tier 2 is built to keep checking rather than assume.

\section{Monitoring: What Aggregates Hide}
\label{sec:tier2}

A router that only reports an aggregate rate -- ``12\% of items went to the slow path'' -- can look healthy while making a systematically bad trade underneath. Tier 2 adds two checks on top of the routing decision itself, logged per item via a decision logger that records the estimates a router acted on, the resulting path and cost, and (because this is a simulation) the hidden ground truth the router never saw.

\subsection{Spend concentration by category}

Table~\ref{tab:spend} aggregates one decision log by category, comparing each category's share of routing spend against its share of traffic. \texttt{premium} and \texttt{luxury} together are 10.9\% of volume but absorb 67.7\% of total spend -- by design, since every item in both categories clears the value threshold. This is the router doing exactly what Section~\ref{sec:tier1} argued it should: concentrating expensive reasoning where a mistake is costly, not where volume is highest. Running the same aggregation over the \textsc{Random} baseline instead yields spend that tracks volume almost exactly, with no segment favored or starved -- the intended contrast, since random routing has no opinion about value at all.

\begin{table}[t]
  \centering
  \caption{Spend by category vs.\ volume share, \textsc{ValueWeighted}, $n{=}2{,}000$.}
  \label{tab:spend}
  \small
  \begin{tabular}{lrrrl}
    \toprule
    Category & Vol.\ \% & Slow \% & Spend \% & \\
    \midrule
    commodity & 39.5 & 0.0   & 12.1 & under \\
    accessory & 32.4 & 0.0   & 10.0 & under \\
    mid\_tier & 17.2 & 4.9   & 10.3 & under \\
    premium   & 7.8  & 100.0 & 48.3 & over \\
    luxury    & 3.1  & 100.0 & 19.4 & over \\
    \bottomrule
  \end{tabular}
\end{table}

\subsection{Calibration collapses within category}

Overall, both estimators look well calibrated against ground truth: value has mean absolute error 3.90 (23.0\% mean absolute \% error) and correlation 0.989; difficulty has mean absolute error 0.10 (37.1\%) and correlation 0.792. Restricting the same correlation check to items \textit{within} each category tells a different story (Table~\ref{tab:calib}).

\begin{table}[t]
  \centering
  \caption{Estimate-vs-truth correlation, computed within each category.}
  \label{tab:calib}
  \small
  \begin{tabular}{lrr}
    \toprule
    Category & Value corr.\ & Difficulty corr.\ \\
    \midrule
    commodity & 0.744 & $-$0.023 \\
    accessory & 0.805 & $-$0.025 \\
    mid\_tier & 0.843 & $-$0.075 \\
    premium   & 0.907 &  0.013 \\
    luxury    & 0.938 & $-$0.047 \\
    \bottomrule
  \end{tabular}
\end{table}

The value estimator's within-category correlation stays high (0.744--0.938): it is genuinely distinguishing higher- and lower-value items \textit{inside} a single category, not merely riding the category average. The difficulty estimator's within-category correlation collapses to approximately zero everywhere. This is the paper's most important methodological finding: categories differ hugely in typical difficulty range, so an estimator that does nothing more than encode which category an item belongs to will already earn a healthy-looking \textit{overall} correlation, purely from separating categories correctly -- not from saying anything true about any individual item. The overall 0.792 number was almost entirely a between-category effect; conditioned on category, the difficulty scorer has essentially no power to tell a hard item from an easy one. An aggregate metric could not have surfaced this; a segment-level breakdown cannot hide it. We consider this the central practical lesson of Tier 2: \textit{do not trust an aggregate calibration number until it has been broken down by the segments the system actually has to discriminate within.}

\section{Seasonal Robustness}
\label{sec:tier3}

Tiers 1--2 evaluate a router against one static batch of traffic. Real deployments face periods where the traffic mix itself shifts -- a seasonal demand surge being the canonical example. Tier 3 asks whether a router (and its budget) tuned on ``normal'' traffic quietly breaks down exactly when it matters most, and what a robust design should look like instead. We stress at the outset that this is a toy demonstration: the ``peak'' period below is a hand-picked multiplier, not a calibration against any real multi-year seasonal dataset.

\subsection{Simulating a demand surge}

The simulator is extended to generate a period of days rather than one flat batch. Each day is tagged \texttt{normal} or \texttt{peak}; a \texttt{peak} day (a stand-in for a Black-Friday-style event) applies a 2.5$\times$ overall volume multiplier together with a category weight shift toward gift-buying categories (\texttt{premium} $\times2.0$, \texttt{luxury} $\times2.5$, \texttt{commodity} $\times0.7$). Table~\ref{tab:seasonmix} shows the resulting category mix shift. Critically, the difficulty scorer and value estimator are \textit{not} recalibrated for peak days -- the same fixed category priors run every day, so any robustness observed below comes from the routing/budget policy, not from the estimators knowing a season shift occurred.

\begin{table}[t]
  \centering
  \caption{Category mix, normal vs.\ peak days (30-day period, 3 peak days).}
  \label{tab:seasonmix}
  \small
  \begin{tabular}{lrr}
    \toprule
    Category & Normal \% & Peak \% \\
    \midrule
    commodity & 39.7 & 28.5 \\
    accessory & 30.8 & 28.1 \\
    mid\_tier & 16.8 & 20.4 \\
    premium   & 9.3  & 15.5 \\
    luxury    & 3.5  & 7.5  \\
    \bottomrule
  \end{tabular}
\end{table}

\subsection{Policies compared}

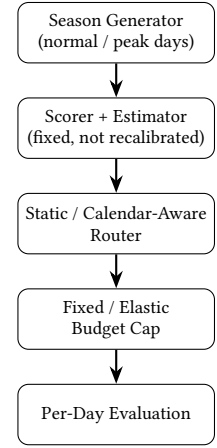
\begin{figure}[t]
  \centering
  \begin{tikzpicture}[
    box/.style={draw, rectangle, rounded corners, align=center, minimum height=0.8cm, minimum width=2.6cm, font=\footnotesize, inner sep=2pt},
    arrow/.style={-{Stealth}, thick},
    node distance=0.45cm
  ]
    \node[box] (season) {Season Generator\\(normal / peak days)};
    \node[box, below=of season] (est2) {Scorer + Estimator\\(fixed, not recalibrated)};
    \node[box, below=of est2] (pol) {Static / Calendar-Aware\\Router};
    \node[box, below=of pol] (cap) {Fixed / Elastic\\Budget Cap};
    \node[box, below=of cap] (eval2) {Per-Day Evaluation};

    \draw[arrow] (season) -- (est2);
    \draw[arrow] (est2) -- (pol);
    \draw[arrow] (pol) -- (cap);
    \draw[arrow] (cap) -- (eval2);
  \end{tikzpicture}
  \caption{Tier 3 evaluation loop. Each day is scored and routed independently; a policy is applied per day and graded before aggregating into period totals.}
  \label{fig:season}
\end{figure}

Four policies are evaluated day-by-day over the period, all acting only on estimates, following the evaluation loop in Figure~\ref{fig:season}:

\textbf{Static}: the Section~\ref{sec:tier1} router with one fixed $\tau_v$, season-blind.

\textbf{Calendar-aware}: the same router, but $\tau_v$ is scaled by a season-specific multiplier (Listing~\ref{lst:calendar}), representing an ops team manually tightening the value bar for a known high-value period.

\begin{lstlisting}[caption={Calendar-aware threshold scaling.}, label={lst:calendar}]
def route(self, d_est, v_est, season="normal"):
    mult = self.season_multipliers.get(season, 1.0)
    threshold = self.value_threshold * mult
    is_hard = d_est >= self.difficulty_threshold
    is_valuable = v_est >= threshold
    return "slow" if (is_hard and is_valuable) else "fast"
\end{lstlisting}

\textbf{Elastic budget}: the static router's raw decisions, with the slow path capped at $\alpha$ times \textit{that day's own} total estimated value (Listing~\ref{lst:elastic}) -- never ground truth, and with no per-season logic at all.

\begin{lstlisting}[caption={Elastic budget cap.}, label={lst:elastic}]
def apply_elastic_cap(decisions, alpha, slow_cost=20.0):
    forecast = sum(d.value_estimate for d in decisions)
    cap = alpha * forecast
    # admit slow-path candidates by value_estimate,
    # highest first, until running cost exceeds cap
    ...
\end{lstlisting}

\textbf{Fixed budget}: the naive alternative the elastic policy is meant to improve on -- one constant daily cap, calibrated off normal-day spend, never adjusted for the day's own value mix.

\begin{equation}
  \text{cap}_{\text{elastic}}(t) = \alpha \sum_{i \in \text{day } t} \widehat{v}_i, \qquad
  \text{cap}_{\text{fixed}}(t) = k
  \label{eq:budget}
\end{equation}

\subsection{Results}

Table~\ref{tab:peak} reports peak-day-only aggregates over a 30-day period with 3 peak days ($\alpha{=}0.08$, fixed cap $k{=}540$, calendar multiplier 2.5$\times$); Table~\ref{tab:normal} reports the same aggregates restricted to normal days, to confirm no policy regresses on ordinary traffic.

\begin{table}[t]
  \centering
  \caption{Peak days only. Budget/value in unit-cost / value units.}
  \label{tab:peak}
  \small
  \begin{tabular}{lrrrr}
    \toprule
    Policy & Slow \% & Budget & Recall & Efficiency \\
    \midrule
    Static          & 23.5 & 8{,}188 & 70.1\% & 10.999 \\
    Calendar-aware  & 22.3 & 7{,}846 & 66.7\% & 11.392 \\
    Elastic budget  & 23.5 & 8{,}188 & 70.1\% & 10.999 \\
    Fixed budget    & 5.4  & 3{,}039 & 16.2\% & 18.901 \\
    \bottomrule
  \end{tabular}
\end{table}

\begin{table}[t]
  \centering
  \caption{Normal days only, same period and settings.}
  \label{tab:normal}
  \small
  \begin{tabular}{lrrrr}
    \toprule
    Policy & Slow \% & Budget & Recall & Efficiency \\
    \midrule
    Static          & 13.5 & 19{,}213 & 63.6\% & 8.421 \\
    Calendar-aware  & 13.5 & 19{,}213 & 63.6\% & 8.421 \\
    Elastic budget  & 12.4 & 18{,}111 & 58.9\% & 8.777 \\
    Fixed budget    & 12.4 & 18{,}111 & 58.8\% & 8.790 \\
    \bottomrule
  \end{tabular}
\end{table}

Three findings stand out. First, the fixed budget collapses on peak days: sized for normal-day spend, it can only afford a fraction of what the surge actually needs, and high-value recall falls from 70.1\% to 16.2\% -- a drop no other policy comes close to. Second, the elastic budget does not merely cope with the surge, it does not notice it: its cap grows with the day's own estimated value, so at these settings it never actually binds on peak days, and its peak-day numbers are \textit{identical} to the unconstrained static router. It absorbed a 2.5$\times$ volume surge and a category mix shift using a rule that has no explicit knowledge that either happened. Third, calendar-aware -- despite being the only policy explicitly told the season -- does worse on the recall/spend trade-off than the elastic budget: it saves some spend (7{,}846 vs.\ 8{,}188) but gives up more recall (66.7\% vs.\ 70.1\%) than the elastic budget ever had to give up. On normal days, calendar-aware is identical to static by construction (its peak multiplier is 1.0 when the season is \texttt{normal}), and elastic and fixed budget land in nearly the same place, since the fixed cap was deliberately calibrated off normal-day spend -- exactly the day type it was built for.

\subsection{Why the elastic cap does not need to know the calendar}

The mechanism is simpler than it first appears. Whether a budget cap binds is governed by the \textit{ratio} of slow-path cost needed to that day's total estimated value, not by the absolute size of either quantity. In this simulation, that ratio is similar on peak and normal days, because cost and value scale up together when volume surges -- a cap set as a fixed fraction of estimated value therefore ends up correctly sized on both kinds of days automatically. The elastic policy is not season-aware; it is simply measuring the right quantity (value estimates the router already has to compute) instead of a dollar figure fixed in advance and never revisited.

\section{Discussion and Limitations}
\label{sec:discussion}

Every result above comes from a controlled synthetic environment where value and difficulty are defined by construction and the ``seasonal surge'' is a hand-picked multiplier rather than a fit to historical data. This is a deliberate choice -- it is what makes the calibration-collapse finding in Section~\ref{sec:tier2} and the elastic-budget mechanism in Section~\ref{sec:tier3} demonstrable at all, since a router can only be graded against ground truth the experimenter is willing to hide from it -- but it also means the specific numbers reported (60\% recall, a 2.5$\times$ surge, an $\alpha$ of 0.08) are properties of this particular synthetic configuration and should not be read as calibrated estimates of what a production system would see. Several limitations follow directly from this:

\begin{itemize}
  \item \textbf{Single-seed runs.} Each table reflects one random seed per configuration; we have not characterized variance across seeds or performed significance testing, which would be necessary before treating any comparison here as more than illustrative.
  \item \textbf{Fixed, non-learned thresholds.} Every router in this study is a hand-set threshold rule. A learned or adaptive router -- in the spirit of the cascade and routing literature in Section~\ref{sec:related} -- could plausibly do better, at the cost of needing labeled outcome data this synthetic environment does not have to offer realistically.
  \item \textbf{No cross-item reasoning.} The \textit{bundle} merchandising action implicitly requires choosing a companion item from the rest of the catalog, but nothing in the current pipeline selects that companion -- the action can be recommended without being fully executable, a gap noted during development as the natural next extension.
  \item \textbf{Synthetic seasonality.} The peak-day multipliers in Section~\ref{sec:tier3} encode a plausible story about holiday shopping, not a fit to any real retailer's historical data. The qualitative mechanism (elastic budgets absorb surges that fixed budgets cannot) is the intended takeaway, not the specific 70.1\%-to-16.2\% recall drop.
\end{itemize}

Future work in roughly the order it would matter for turning this into something closer to a validated system: multi-seed evaluation with confidence intervals; a learned router or budget policy trained against logged outcomes rather than hand-set thresholds; calibration against a real (even if small) e-commerce dataset in place of the synthetic simulator; and closing the \textit{bundle} gap with an explicit companion-item selection step.

\section{Implications for LLM Inference Cost Optimization}
\label{sec:implications}

Although the pipeline in this paper is framed as a retail merchandising simulation, the three findings translate directly to any system routing between a cheap and an expensive LLM-backed path -- a confidence-based cascade, a small-model/large-model router, or a gate deciding whether to invoke an expensive agentic loop or tool call. We make that translation explicit here for readers working on LLM serving-cost optimization rather than e-commerce specifically.

\textbf{Add a value signal alongside the confidence signal.} Most deployed cascades and routers (Section~\ref{sec:related}) escalate to an expensive model based on some notion of the cheap model's uncertainty or predicted quality -- the LLM-serving analogue of our difficulty estimate. Section~\ref{sec:tier1results} shows that weighting the same escalation decision by value, not just uncertainty, improves precision without giving up recall: an uncertainty-only router spends its expensive-model budget equally on high-stakes and low-stakes uncertain queries, while a router that also weighs \textit{what a wrong answer would cost} (a support reply that could cause churn vs.\ one that won't; a code-generation request touching production infrastructure vs.\ a scratch script) concentrates escalation exactly where it pays for itself. Wherever uncertainty is already estimated for cascade routing, adding a cheap, coarse value or stakes estimate is a small addition with a measurable precision payoff.

\textbf{Calibrate confidence estimators per segment, not only in aggregate.} Section~\ref{sec:tier2} is the more general methodological point, and arguably the one most transferable outside this simulation: an aggregate correlation between a router's confidence score and actual correctness can look healthy purely because the estimator has learned to separate easy \textit{query types} from hard ones, while carrying no real discriminative power \textit{within} a query type. Anyone validating a confidence-based router or cascade should re-run the same calibration check conditioned on query category, user segment, or task type before trusting an overall number -- exactly the failure our difficulty scorer exhibited (0.792 correlation overall, collapsing to $\approx$0 within every category in Table~\ref{tab:calib}).

\textbf{Let cost budgets scale with estimated demand, not a fixed figure.} Section~\ref{sec:tier3}'s elastic-budget result applies directly to any team operating an LLM API budget under variable load -- a traffic spike, a seasonal peak, a feature launch that changes usage patterns overnight. A fixed daily or monthly spend cap sized for typical traffic will either throttle legitimate expensive-path requests during a surge (our fixed-budget policy's high-value recall collapse, 70.1\%~$\to$~16.2\%, Table~\ref{tab:peak}) or sit needlessly conservative on quiet days. A cap that instead scales with a cheap, already-computed signal -- e.g., the router's own aggregate confidence or value estimates for the current batch of traffic -- adapts to load the system is already measuring internally, without needing to know in advance that a surge is happening.

\section{Conclusion}

Across all three tiers, the same pattern recurs: a signal the system already has to compute -- a value estimate, a per-category breakdown of an aggregate number, a day's own running total -- turns out to carry more useful information than a rule fixed in advance, whether that rule is a difficulty-only threshold, an unbroken-down calibration metric, or a flat dollar budget. Value-weighted routing beats difficulty-only routing on precision by using a signal the difficulty-only policy already had available but ignored. The calibration monitor in Section~\ref{sec:tier2} catches a difficulty estimator that looks calibrated in aggregate only because it has learned to distinguish categories, not items. The elastic budget in Section~\ref{sec:tier3} absorbs an unannounced demand surge by reacting to its own estimates rather than to a number someone chose once and forgot to revisit. None of these findings required a more sophisticated model -- each came from asking a more specific question of the same underlying signals the simpler policy already had.

\section*{Availability}

Code, CLI entry points, and default configuration for every result in this paper are available at \url{https://github.com/BhavtoshRath/value-router}~\cite{valuerouter2026}. In particular: Table~\ref{tab:catalog} and Table~\ref{tab:tier1} reproduce via \texttt{python -m value\_router.eval\_harness -n 2000 --seed 42}; Table~\ref{tab:spend} and Table~\ref{tab:calib} via \texttt{python -m value\_router.decision\_logger} followed by \texttt{python -m value\_router.monitor --log decision\_log.jsonl}; and Tables~\ref{tab:seasonmix}, \ref{tab:peak}, \ref{tab:normal} via \texttt{python -m value\_router.season\_eval --n-days 30 --peak-days 25,26,27}.

\bibliographystyle{ACM-Reference-Format}
\bibliography{references}

@article{chen2023frugalgpt,
  title   = {FrugalGPT: How to Use Large Language Models While Reducing Cost and Improving Performance},
  author  = {Chen, Lingjiao and Zaharia, Matei and Zou, James},
  journal = {arXiv preprint arXiv:2305.05176},
  year    = {2023}
}

@article{yue2023cascades,
  title   = {Large Language Model Cascades with Mixture of Thought Representations for Cost-Efficient Reasoning},
  author  = {Yue, Murong and Zhao, Jie and Zhang, Min and Du, Liang and Yao, Ziyu},
  journal = {arXiv preprint arXiv:2310.03094},
  year    = {2023}
}

@article{madaan2023automix,
  title   = {AutoMix: Automatically Mixing Language Models},
  author  = {Madaan, Aman and Aggarwal, Pranjal and Anand, Ankit and Potharaju, Srividya Pranavi and Mishra, Swaroop and Zhou, Pei and Gupta, Aditya and Rajagopal, Dheeraj and Kappaganthu, Krishna and Yang, Yiming and others},
  journal = {arXiv preprint arXiv:2310.12963},
  year    = {2023}
}

@article{ong2024routellm,
  title   = {RouteLLM: Learning to Route LLMs with Preference Data},
  author  = {Ong, Isaac and Almahairi, Amjad and Wu, Vincent and Chiang, Wei-Lin and Wu, Tianhao and Gonzalez, Joseph E. and Kadous, M. Waleed and Stoica, Ion},
  journal = {arXiv preprint arXiv:2406.18665},
  year    = {2024}
}

@misc{valuerouter2026,
  title        = {value-router: Simulating Value-Weighted Routing Decisions over a Stream of Items},
  author       = {Rath, Bhavtosh},
  year         = {2026},
  howpublished = {\url{https://bhavtoshrath.github.io/posts/value-router-value-weighted-routing}},
  note         = {Blog post}
}

@inproceedings{chen2020frugalml,
  title={{FrugalML}: How to Use {ML} Prediction {API}s More Accurately and Cheaply},
  author={Chen, Lingjiao and Zaharia, Matei and Zou, James},
  booktitle={Advances in Neural Information Processing Systems (NeurIPS)},
  volume={33},
  pages={10685--10696},
  year={2020}
}

@inproceedings{wang2023tabi,
  title={Tabi: An Efficient Multi-Level Inference System for Large Language Models},
  author={Wang, Yiding and Chen, Kai and Tan, Haisheng and Guo, Kun},
  booktitle={Proceedings of the Eighteenth European Conference on Computer Systems (EuroSys)},
  pages={233--248},
  year={2023}
}

@inproceedings{ding2024hybrid,
  title={Hybrid {LLM}: Cost-Efficient and Quality-Aware Query Routing},
  author={Ding, Dujian and Mallick, Ankur and Wang, Chi and Sim, Robert and Mukherjee, Subhabrata and Ruhle, Victor and Lakshmanan, Laks V.S. and Awadallah, Ahmed Hassan},
  booktitle={International Conference on Learning Representations (ICLR)},
  year={2024}
}

@article{zhang2023ecoassistant,
  title={{EcoAssistant}: Using {LLM} Assistant More Affordably and Accurately},
  author={Zhang, Jieyu and Krishna, Ranjay and Awadallah, Ahmed H. and Wang, Chi},
  journal={arXiv preprint arXiv:2310.03046},
  year={2023}
}

@inproceedings{varshney2022model,
  title={Model Cascading: Towards Jointly Improving Efficiency and Accuracy of {NLP} Systems},
  author={Varshney, Neeraj and Baral, Chitta},
  booktitle={Proceedings of the 2022 Conference on Empirical Methods in Natural Language Processing (EMNLP)},
  pages={11007--11021},
  year={2022}
}

@inproceedings{guo2017calibration,
  title={On Calibration of Modern Neural Networks},
  author={Guo, Chuan and Pleiss, Geoff and Sun, Yu and Weinberger, Kilian Q.},
  booktitle={Proceedings of the 34th International Conference on Machine Learning (ICML)},
  pages={1321--1330},
  year={2017}
}

@article{kadavath2022language,
  title={Language Models (Mostly) Know What They Know},
  author={Kadavath, Saurav and Conerly, Tom and Askell, Amanda and Henighan, Tom and Drain, Dawn and Perez, Ethan and Schiefer, Nicholas and Dodds, Zac Hatfield and DasSarma, Nova and Tran-Johnson, Eli and others},
  journal={arXiv preprint arXiv:2207.05221},
  year={2022}
}

@inproceedings{tian2023just,
  title={Just Ask for Calibration: Strategies for Eliciting Calibrated Confidence Scores from Language Models Fine-Tuned with Human Feedback},
  author={Tian, Katherine and Mitchell, Eric and Zhou, Allan and Sharma, Archit and Rafailov, Rafael and Yao, Huaxiu and Finn, Chelsea and Manning, Christopher D.},
  booktitle={Proceedings of the 2023 Conference on Empirical Methods in Natural Language Processing (EMNLP)},
  pages={5433--5442},
  year={2023}
}

@inproceedings{elkan2001foundations,
  title={The Foundations of Cost-Sensitive Learning},
  author={Elkan, Charles},
  booktitle={Proceedings of the 17th International Joint Conference on Artificial Intelligence (IJCAI)},
  volume={2},
  pages={973--978},
  year={2001}
}

@inproceedings{leviathan2023fast,
  title={Fast Inference from Transformers via Speculative Decoding},
  author={Leviathan, Yaniv and Kalman, Matan and Matias, Yossi},
  booktitle={Proceedings of the 40th International Conference on Machine Learning (ICML)},
  pages={19274--19286},
  year={2023}
}

@article{fedus2022switch,
  title={Switch Transformers: Scaling to Trillion Parameter Models with Simple and Efficient Sparsity},
  author={Fedus, William and Zoph, Barret and Shazeer, Noam},
  journal={Journal of Machine Learning Research},
  volume={23},
  number={120},
  pages={1--39},
  year={2022}
}

\end{document}